\documentclass{article}
\usepackage{spconf,amsmath,graphicx,amssymb}

\usepackage{xspace}
\def\etal{{\em et al}\@\xspace}
\def\ie{i.e.\@\xspace}
\def\eg{e.g.\@\xspace}

\usepackage{multirow}
\usepackage{pifont}

\title{DP-Net: Learning Discriminative Parts for image recognition}
\name{R. Sicre, H. Zhang, J. Dejasmin, C. Daaloul, S. Ayache, T. Artières%\thanks{Thanks to XYZ agency for funding.}
}
\address{Centrale Marseille, Aix Marseille Univ, CNRS, LIS, Marseille, France}
\begin{document}
%\ninept
%

Copyright 2023 IEEE. Published in 2023 IEEE International Conference on Image Processing (ICIP), scheduled for 8-11 October 2023 in Kuala Lumpur, Malaysia. Personal use of this material is permitted. However, permission to reprint/republish this material for advertising or promotional purposes or for creating new collective works for resale or redistribution to servers or lists, or to reuse any copyrighted component of this work in other works, must be obtained from the IEEE. Contact: Manager, Copyrights and Permissions / IEEE Service Center / 445 Hoes Lane / P.O. Box 1331 / Piscataway, NJ 08855-1331, USA. Telephone: + Intl. 908-562-3966.

\newpage

\maketitle
\begin{abstract}

This paper\footnote{This work has received funding from the UnLIR ANR project (ANR-19-CE23-0009), the Excellence Initiative of Aix-Marseille Universite - A*Midex, a French “Investissements d’Avenir programme” (AMX-21-IET-017). Part of this work was performed using HPC resources from GENCI-IDRIS (Grant 2020-AD011011853 and 2020-AD011013110).} presents Discriminative Part Network (DP-Net), a deep architecture with strong interpretation capabilities, %that address image classification problems.
% such as fine-grained classification.
%The approach 
which exploits a pretrained Convolutional Neural Network (CNN) combined with a part-based recognition module.
This system learns and detects parts in the images that are discriminative among categories, without the need for fine-tuning the CNN, making it more scalable than other part-based models.
While part-based approaches naturally offer interpretable representations, 
we propose explanations at image and category levels and introduce specific constraints on the part learning process to make them more discrimative.
%Our experiments show that representations based on parts can outperform global representations on some datasets.%, MIT-67 Scenes, CUB-200-2011, and ImageNet, using two standard architectures, VGG and ResNet.

%%%%%%%%%%%%%%%%%%%%%%\keywords{image classification, part-based models, interpretability.}
\end{abstract}


\section{Introduction}

Since 2012, Deep Neural Networks (DNN) have been re-popularized in the fields of computer vision and machine learning.
Deep learning methods are used to address almost every single computer vision problem such as image classification, retrieval, detection, segmentation, etc.
The ability to transfer pre-trained representations learned on large annotated datasets led to large improvements.
%Furthermore, architectures are adapted to further improve the learning and adaptation process.

With most efforts dedicated to improving further these methods, it is interesting to remember previous methods and concepts that occur prior to the deep learning tsunami.
For instance, a large effort has been dedicated to Part-Based Models (PBM) starting with the deformable part model \cite{felzenszwalb10}.
Based on these models, diverse strategies were later proposed to learn a collection of discriminative parts \cite{juneja2013blocks,doersch13}.
Global image representations are further derived from the learned parts to perform recognition, while mainstream  methods would aggregate dense local representations instead \cite{perronnin10}.

With the re-popularization of deep learning, links between part-based models and CNN architectures are discussed in \cite{girshick2015deformable}.
First, PBM started using pre-trained network to replace previous feature extraction techniques \cite{Mettes15,parizi2014automatic,Sicre15A}.
%, maintaining their superiority over standard architecture \cite{Mettes15,parizi2014automatic,Sicre15A}
%, based on the aggregation of dense representations \cite{perronnin10}.
%
%
Later, part inspired architectures are introduced to target the problem of fine-grained classification \cite{tt15,tt24,tt26} or scene recognition \cite{chen2019scene}, showing the benefit of such architectures on specific datasets and tasks.
Furthermore, ProtoPNet \cite{thisthat} and its extensions \cite{rymarczyk21,nauta21,rymarczyk22} provide new interpretability ability by exploiting relations between categories and parts.
However, these architectures still need multiple training stages; can not cope with large scale dataset as ImageNet \cite{deng09}; require extra annotations such as bounding boxes or part annotations; or do not provide more interpretability than part-level visualization.

Our method is a part-based architecture that addresses these limitations.
Learning discriminative parts is performed jointly with the classification. 
Unlike previous models, we exploit a pretrained DNN to compute representations on a number of random regions in images, in order to improve simplicity and scalability.
Parts are learned solely based on image level labels, \ie without any priors or bounding boxes. 
Moreover, our model provides explanations of parts and classification decision, for a specific input or more broadly for a category. 
Finally, we show that introducing a number of constraints on the parts in the objective function helps discovering more discriminative and relevant parts.
%

%------------------------------------------------------------------------------
\begin{figure*}[h]
\centering
\includegraphics[width =.9 \textwidth]{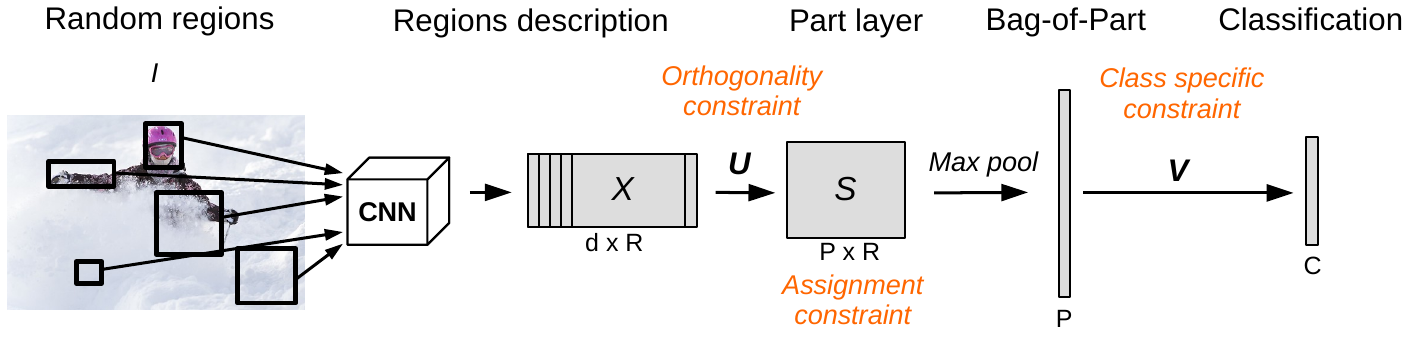} \ 
\caption{Figure of the proposed architecture and its learned parameters $U$ and $V$.
%Pourquoi Conv1D??? Ajouter les contraintes dans le schéma? 
}
\label{fig1}
\end{figure*}
%---------------

\section{Previous works}

\paragraph*{Early part-based approaches}

Most approaches define parts as image regions that can help differentiate between the categories.
However, methods vary in how they select these regions.
The constellation model \cite{weber10} and deformable part model \cite{felzenszwalb10} are the original methods that model classes by parts and their positions.
%
%Inspired by \cite{Ullman2001gh}, l
Later approaches \cite{juneja2013blocks,doersch13,Sicre16A,sicre17,Mettes15}  first learn the parts, then the decision function, based on parts response. 
Parts are learned by identifying candidate regions \cite{juneja2013blocks}, using mean-shift algorithm \cite{doersch13}, an iterative \emph{softassign} matching algorithm \cite{Sicre16A,sicre17}, or boosting \cite{Mettes15}.
%%%%%%%For instance Juneja \etal \cite{juneja2013blocks} iteratively identifies candidate regions and learn part classifiers. Doersch \etal \cite{doersch13} discovers discriminative regions with a density-based mean-shift algorithm. Sicre \etal \cite{sicre17} use a \emph{softassign}-like matching algorithm to iteratively refine parts and match regions to parts. Mettes \etal \cite{Mettes15} learns parts with a boosting approaches that are shared across categories.
%
However Parizi \etal \cite{parizi2014automatic} jointly learns the parts and the category classifiers.
%Parts are explicitly optimized for the final classification.
%The parts are represented as a combination of regions and are explicitly optimized for the final classification.

%Our work has a similar objective to these part-based methods. 
Our method also optimizes the final classification in a single-stage optimization. 
We further study constraints on parts inspired by these works \cite{Mettes15,sicre17}.

\paragraph*{Part-inspired architecture}

Several recent part inspired methods focus on fine-grained image classification.
Most of these methods follow detection architectures, as R-CNN \cite{rcnn,ren15}, where detection and classification are learned alternatively.
Other works are based on attention models \cite{tt49,tt26,tt7}. 
Recently, Chen \etal \cite{chen2019scene} propose to build protoype-agnostic scene layout, using graph convolutions to encode parts and their relations. 
Also, Krause \etal \cite{tt19} generate parts using co-segmentation and alignment. 
%Neural activation constellations \cite{SiRo15} seeks constellation pattern in convolution filters.

ProtoPNet \cite{thisthat} propose to learn parts prototype in three stages, which is soon improved by allowing shared parts \cite{rymarczyk21}, negative parts \cite{nauta21} and more effective optimization \cite{rymarczyk22}.
These recent methods are directly aligned with our objectives, but their optimization remain complex with finetuning of the backbone network, fixed size region, pruning requirements, etc.
Our method thus offer a simpler adaptable architecture that can learn on large datasets.

\paragraph*{Interpretability}

Interpretability recently receives a lot attention \cite{lipton2016mythos}. Numerous methods consider models as black boxes, and perform post-hoc interpretability.
While other methods aim at changing models, or representations, to make them more transparent or easily understandable.

Some transparency can be obtained through parts.
However, these methods can not easily link parts representation with the decision, except for ProtoPNet \cite{thisthat}, which is the first work to propose visualizable parts and their contributions to a decision at inference.

%Interestingly, CAM \cite{cam} offers post-hoc interpretability regarding the decision, based on feature activations.
%And ProtoPNet \cite{thisthat} is the first work to propose visualizable parts and their contribution to a decision at inference.%a part based model that offer explainability at inference by highlighting the regions that contribute the most in the decision, as well as linking them to regions from the training set. 
%

In our work, we also provide explainability at inference, but at a lower computational cost. 
We also extend the posthoc method CAM \cite{cam} to part models, enabling various levels of interpretability. 
Finally, we study various constraints to provide both discriminative and interpretable parts. 
%%%

%\textcolor{red}{on peut dire ca ? qu'on adapte ? ca signifie contribution, et c'est mioeux, meme si c'est tres leger}.

%\textcolor{red}{[2] n'utilise pas CAM!}

\section{Discriminative Parts NN (DP-Net)}

%We first present our architecture, then discuss the learning and constraints on the part.

%DP-net is specifically designed to perform accurate and interpretable classification for complex recognition tasks such as fine-grained classification. 
%DP-nets core is a part-based model that are known to be particularly efficient on such problems and can provide interpretable features\cite{thisthat}.
%
%After a description of the architecture and the learning process, we show our constraint definition to improve part interpretability.

\subsection{DP-net architecture} 

In a nutshell, the DP-net is a part-based architecture that includes: 1) a pretrained CNN, \ie backbone, that produces high-level descriptions of numerous (randomly selected) regions in an input image, 2) a part layer that outputs a matching score for every part-region pair, 3) a max-pooling that encodes parts information in a dense vector representation, and 4) a final classification layer, see Fig. \ref{fig1}. Using a pretrained CNN (that is not refined) and randomly selected regions makes our method particularly simple, adaptable and scalable.

In more details, when processing an input image $I \in \mathcal{I}$, a fixed number $R$ of rectangular regions, $\mathcal{R} = \{i_1,..., i_R\}$, are extracted and then described by a $d-$dimensional descriptor vector output by a pretrained CNN, %(by average pooling over the outputs of a pretrained CNN 
\eg VGG19, ResNet50, etc.
The resulting matrix is defined as $X \in \mathbb{R}^{d \times R}$,
whose $r^{th}$ column is the descriptor of the $r^{th}$ region, $x^r=CNN(i^r)\in\mathbb{R}^d$.

The part layer computes a matching score between regions and parts that are represented as $d-$dimensional vectors living in the same space as region descriptors.
We note $U \in \mathbb{R}^{P \times d}$ the part layer weights, with $u_p$ the representation of the $p^{th}$ part.
The output of this layer is defined as the score matrix $S \in \mathbb{R}^{P \times R}$ with $S = U \times X =  (s_{p,r})_{p,r \in [1,...P] \times [1,...R]} $ with $s_{p,r} =  u_p \times x^{r} $.
This score matrix is often referred to as the assignment matrix as it relates extracted regions to the learned parts.
We further apply one-dimensional max-pooling, and $L2$ normalization, to obtain a final "bag-of-parts" (BOP) embedding, introduced in \cite{juneja2013blocks} and used in most part-based method.
The bag-of-parts is a $P$-dimensional vector given as input to a final classification layer, \ie a fully connected layer, with a softmax activation function, with weights $V \in \mathbb{R}^{C \times P}$ where $C$ denotes the number of classes. To summarize, for an input image $I$, the model computes successively $X$, $S$, $b$ and $o$, as:

\begin{center}
\begin{tabular}{cc}
%\begin{equation}
$X = [CNN(i_1(I)), ..., CNN(i_R(I))]$, \ \ \ \ $S =  U \times X$, \\

$b = \left[\max\limits_{r \in R} (s_{p,r})\right]_ {p \in [1,...P] }$, \ \ \ \ \ \ \ \ \ \ \ \  $o =  \text{softmax}(V \times b)$. \\
\end{tabular}
\end{center}
%\end{equation}

\subsection{Learning }
\label{sec:constraints}
Our model simply learns the part matrix $U$ and the classification layer $V$, through gradient descent, to minimize categorical cross-entropy noted as $C_{CCE}(U,V)$. 

Actually the most important aspect of the learning lies in constraints that are added through additional terms to the objective function, to improve interpretability.
Here is a list of desirable properties of the parts:
%\begin{comment}
%    \item Parts should be complementary or diverse to provide more information, 
%    \item Parts should cover as much as possible the many regions one may extract from an image,
%    \item Parts should be discriminative with respect to class labels,
%    \item To favour interpretability the parts should be specific to classes.
%\end{comment}

%\vspace{0.2cm}
1) Parts should be complementary, \ie parts should be different one from another.

2) Parts should cover as much as possible the diversity of regions extracted from images.

3) Parts should be discriminative with respect to classes.

4) Parts should be specific to categories.
%\vspace{0.2cm}

%Property 1) and 2) should maximize the information brought by parts;
%property 3) maximizes accuracy and interpretability; while property 4) should help interpretability.
These properties can be satisfied by including additional terms in the objective function. 
We note that property 3) is already addressed by $C_{CCE}(U,V)$.

Specifically, the first property is induced by enforcing parts to be \emph{orthogonal} one to another, simply adding to the objective function a term as $C_{\perp}(U) = - \frac{1}{P^2} \sum\limits_{i=1}^{P} \sum\limits_{j=1, j \neq i}^{P}  (u^T_i u_j)^2$. 

The second property can be rewritten as: we want every region to be, as much as possible, \emph{assigned} to one of the parts, \eg following \cite{sicre17}. 
This can be easily encouraged by minimizing the entropy of each column of the score matrix $S$, matching regions to all parts.
Thus, we define $C_{Assign}(U) = - \sum\limits_{r=1}^{R} \sum\limits_{p=1}^{P} s_{p,r} log(s_{p,r})$. Note that softmax is first applied on the columns of the matrix $S$ and each part vector $u_p$ is assumed to be $l2$-normalized for both $C_{\perp}(U)$ and $C_{Assign}(U)$.

Finally, the last property is enforced by adding a constraint on the parameters of the classification layer $V$, by minimizing the contribution of parts that are not assigned to the given class, as in \cite{thisthat}. Let $q$ be the number of parts learned per class, our \emph{class-specific (CS)} constraint can be computed as:
$CS(V) = \frac{1}{P(C-1)}\sum\limits_{i=1}^{C} \sum\limits_{\ \ j=1, j \notin [q(i-1),qi]\ }^{P}  V_{i,j}$

%are applied by introducing auxiliary classifiers, one per class, that operate on a subset of the bag-of-parts representation $b$.
%Thus, parts become \emph{class-specific} ($CS$) and discriminative. 
%This strategy relates to the adversarial learning idea where auxiliary classifiers are exploited to enforce a representation to encode, or not, a particular information. 
%Here we define the number of parts per class as $K$.
%The $K$ first parts are assigned to class 1, the $K$ next parts to class 2, etc. 
%Each classifier receives a $K$-dimensional vector coming from the bag-of-parts and is trained to recognize samples of one class against all the others.
%The Binary Cross Entropy training losses of all these classifiers, whose parameters are noted $Z = (Z_c)_{c \in [1...C]}$, are added to form an auxiliary loss :  $C_{CS}(U,Z) = \sum\limits_{c=1}^{C} C_{BCE}(Z_c,U)$ 

Finally the learning is performed through the minimization of the combined loss function:

\begin{align}
\begin{split}
\mathcal{C}(U,V, Z) & = 
%\min\limits_{w_p} 
C_{CCE}(U,V) + \lambda_1  C_{\perp}(U) \\
& + \lambda_2  C_{Assign}(U) + \lambda_3 C_{CS} (U,Z)
\end{split}
\end{align}

\subsection{Interpretability strategies}

First of all, the learned part $p$ can be simply characterized by the region $r$, when this region produces the highest match scores $s_{p,r}$ over the entire training set. 

Secondly, we want to quantify the importance of the part $p$ for the detection of a particular class $c$. 
This measurement can be obtained through the classification layers that takes as input the bag-of-parts representation $b$, which consist in a single score per part.
Moreover, the output of the model is computed as $o = softmax(V \times b)$ so that the prediction for class $c$ is $o_c = v_c \times b = \sum_p v_{c,p} * b_p$. The importance of part $p$ when predicting class $c$ for a particular input image can then be measured by $v_{c,p} * b_p$. 

At the class level, \cite{thisthat} proposed to interpret categories by the participation of the learned parts in the decision. We follow this idea and adapt here the popular Class Activation Map (CAM) \cite{cam}.
To measure the global importance of part $p$ to recognize class $c$, we want to compute an average of this quantity over the training set. 
Yet, we found more informative to weigh this quantity by the popularity $f(p)$ of the part among all classes.
This frequency term is similar to the $TF.IDF$ weighting scheme from the seminal Vector Space Model in text information retrieval. We thus compute the discriminative power of part $p$ for class $c$ as:
\begin{equation}
d(p,c) = \sum\limits_{I \in \mathcal{I}, y_I = c}  \frac{v_{c,p}*b_p(I)}{log(f(p))}
\label{dMAC}
\end{equation}
where $f(p)$ measures to which extent part $p$ is frequently detected in images of all classes. $f(p)$ is computed as the number of classes  having $p$ in their most activated parts, measured by $v_{c,p} * b_p$.
Given these statistics, given a class $c$, one can visualize the typical regions that are associated with the top parts maximizing $d(p,c)$, by finding regions maximizing $s_{p,r}$. %from the training set maximize the match with these parts (i.e. maximum match scores).

Finally, given an input image and a class $c$, one can use the presented statistics to infer how the image regions participate to the final classification. 
%%%%%%%%%%????
%Regions are linked to parts that are linked to categories within the DP-net architecture. %, via the mappings.
Specifically, we can first identify the top-N most discriminative parts for the class $c$ according to Eq. (\ref{dMAC}).
Then the top-M regions that activate these parts are selected and can be highlighted to compute a heatmap, to help explaining the decision of the model.

\section{Experiments}

%This section presents the public datasets used, implementation details and results obtained with the proposed method.

\paragraph*{Datasets}

%We evaluate DP-Net on fine-grained classification, scene recognition, and large scale data.
We study three datasets: \textbf{CUB-200-2011} \cite{CUB} fine-grained classification of bird species, the \textbf{MIT 67 Scenes} \cite{Quattoni09} dataset of indoor scenes, and the large \textbf{ImageNet} \cite{deng09} (ILSVRC 2012) dataset.
%
%\textbf{MIT 67 Scenes \cite{Quattoni09}} is composed of 6,700 images from 67 categories of indoor scenes. 
%
%\textbf{CUB-200-2011 \cite{CUB}} is a very popular fine-grained dataset of 200 bird species with 11,788 images. 
%
%\textbf{ImageNet \cite{deng09}}  (ILSVRC 2012) is a large image classification dataset composed of around 1.2 million training images from 1,000 categories.
%The 50,000 validation images are used for testing.
%
Only image-level categories are used and no data augmentation is performed.% and methods are evaluated with accuracy.

%///////////////INAT 2019 1010 categories. train set 265213 images. val set 3030.

\paragraph*{Implementation details}

Each image is resized to $544 \times 544$ then given as input to a CNN: VGG-19 or ResNet-50, pretrained on ImageNet.
From the output of the last convolutional layer, we extract the feature maps covering $R$ random regions and perform average pooling on these features. We set $R=500$ for MIT and CUB and $R=100$ for ImageNet.

To learn on MIT-67 and CUB-200, we perform 40 epochs with Adam optimizer and a learning rate of $10^{-3}$. The learning rate is divided by 10 after each 10 epochs, reaching $10^{-6}$.
Since our input data is large, we build large batches of training data and perform 32 batch-level epochs.
% before moving on to the next batch.
%
Concerning ImageNet, we similarly perform 10 batch-level epochs and 4 epochs with learning rate set to $10^{-3}$.

Concerning constraints, after evaluation, the scaling constants are set to $\lambda_1 = 10^{-2}$, $\lambda_2 = 10^{-3}$, and $\lambda_3 = 10^{-3}$.% ????????.

%------------------------------------------------------------------------------
%\begin{figure}[t]
%\centering
%\includegraphics[width = 0.23\textwidth,height =0.16\textwidth ]{images/nb_parts2.png}
%\includegraphics[width = 0.24\textwidth,height =0.16\textwidth]{images/iNet_size2.png}
%\caption{ Figures showing the performance when varying the number of parts for MIT and Birds, on the left, and varying training data quantity for ImageNet, on the right }
%\label{fig2}
%\end{figure}
%---------------
%------------------------------------------------------------------------------

\begin{table}
\begin{center}
\small
 \begin{tabular}{|l|c|c|c|c|c|}
 \hline
  Method &   \multicolumn{2}{|c|}{ISA parts \cite{sicre17}} & \multicolumn{3}{|c|}{Our DP-Net}\\
 \hline
 Dataset &  \multicolumn{2}{|c|}{MIT} & \multicolumn{3}{|c|}{MIT} \\
 \hline
  Network  &   VGG & Places & VGG & Places & RN50 \\
 \hline
  \hline
 Global & 73.3& 78.5 & 76.2 & 79.8  & 78.1 \\
 Parts & 75.1 & 81.6 & 76.9 & 82.0 & 79.7 \\
  \hline
  \end{tabular}
  \normalsize
\end{center}

\begin{center}
\small
 \begin{tabular}{|l|c|c|c|c|}
  %Method &   \multicolumn{2}{|c|}{ISA parts \cite{sicre17}} & \multicolumn{7}{|c|}{Our DP-Net}\\
 \hline
 Dataset & \multicolumn{2}{|c|}{Birds} & \multicolumn{2}{|c|}{ImageNet} \\
 \hline
  Network  &  VGG & RN50 & VGG & RN50 \\
 \hline
  \hline
 Global &  66.4 & 81.5 & 61.0 & 70.8\\
 Parts &  76.1 & 84.9 & 69.0 & 74.6\\
  \hline
  \end{tabular}
  \normalsize
\end{center}
\caption{Tables comparing our DP-Net with global representations on MIT, Birds, and ImageNet datasets using VGG and ResNet. We note that parts are learned without constraints.}
 \label{tab:main}
\end{table}

%\subsection{Results}

\paragraph*{Model performance}

We first show the benefits of using the proposed DP-Net over global representation on the three datasets and two networks, see Table \ref{tab:main}. We learn 20, resp. 10, parts per class for MIT and CUB, resp. ImageNet. 
The global representation model computes global average pooling after the last convolutional layer of the same network followed by either a single, or two, FC layer.
%Using one FC layer is similar to the DP-Net as the classification layer is applied after the representation layer.
%as the  classification layer is directly after the last representation layer
%Adding a second FC layer with ReLU and dropout allows to reach the same number of parameters as our model.
%, or two FC layers (to obtain the same number of parameters as DP-Net).
The best performing global model is presented, as adding the second layer allows to compare the model with the same number of parameters as DP-Net. 
We observe that part methods significantly outperform global models on most cases.%, with less improvement observed on MIT.
%Improvement is more significant for the Birds and ImageNet datasets than for MIT-67.
%
The method is also compared to the part-based method proposed in \cite{sicre17}. %, using the same regions with a VGG-19 pretrained on ImageNet or a VGG-16 learned on Places-205 dataset.
Our model performs similarly but is simpler and more efficient.

Training on ImageNet is particularly interesting, as none of the previous part-based method can cope with such a large dataset.
However, we note that the original models obtain better scores than DP-Net and our global representation: 71.3\% and 74.9\% accuracy for VGG-19 and ResNet-50.
The performance drop for global representation is explained by the image dimension increase and global average pooling added in the case of VGG. 
%More experiments on ImageNet are presented in Figure \ref{fig2}, using smaller amount of training data, \eg from 5k to 200k images.%, and the same test set.
%As expected the performance decreases as we use less training data and using parts always provides better performances.

%We further evaluate how the number of parts learned by the model impacts the performance of DP-Net, see Figure \ref{fig2}, and we observe a plateau when learning around 3,000 parts.

As mentioned earlier, several constraints on parts are evaluated: orthogonality, assignment between regions and parts, and class specific (CS) parts, see Table \ref{tab:reg}.
These constraints allow better interpretation of parts but there is no significant alteration of the performance using these constraints.

\begin{table}[t]
\begin{center}
\small
 \begin{tabular}{|l|c|c|c|c|c|}
 \hline
 Dataset& \multicolumn{4}{|c|}{Constraints} \\
\hline
  & wo & $\perp$ & Assign & CS\\
 %\multirow{6}{*}{Birds} & \multirow{6}{*}{ResNet} & \xmark & \xmark & \xmark &  \\
 \hline
 Birds & 84.9 & 84.6 & 84.6 &  84.5\\
 MIT & 79.7 & 79.1 & 80.3 & 79.5 \\
% Birds & 78.20//77.46 & 75.65 & 77.32 &  75.30\\
 %MIT & 79.70 & 79.10 & 80.30 & 79.25 \\

  \hline
      & $\perp$+Assign & CS+$\perp$ & CS+Assign & CS+$\perp$+Assign\\
       \hline

 Birds & 85.1 & 84.4 & 84.3 & 85.0\\
  MIT & 80.3 & 78.8 & 79.9  & 80.5\\

 %Birds & 76.93 & 75.49 & 75.11 & 75.13\\
 % MIT & 80.29 & 79.10 & 78.88  & 78.96\\

 % & & \cmark & \xmark & \xmark & \\
  %& & \checkmark & & & \\
  %& & & & & \\
  %& & & & & \\
  %& &\cmark & & & \\

 %Dataset & Net. & wo & \perp & Assign & \perp + Assign & CS & CS + \perp & CS + Assign & CS + \perp + Assign \\%Rep. & Last 
 %\hline

%Birds & \multirow{2}{*}{ResNet} & 78.20//77.46 & 75.65 & 77.32 & 76.93 & 75.30 & 75.49 & 75.11 & 75.13  \\
%MIT-67 &  & 79.70 & 79.10 & 80.30 & 80.29 & 79.25 & 79.10 & 78.88  & 78.96\\

\hline

 \end{tabular}
\end{center}
\vspace{-0.2cm}
\caption{Accuracy obtained with ResNet 50 when using the constraints defined in \ref{sec:constraints} (wo = without any constaint).}
 \label{tab:reg}
\end{table}

%NEW with hanwei's data:
%%% Bird  BOP std  84.93

%
%While orthogonality did not have impact on the performance, sparcity constraint tend to reduce the scores.

%%%%%%%%%%%%%%%%%%%%%%%%%%%%%%%%%%%%%%%%%//////////more FC ?

%\subsubsection{}

\paragraph*{Interpretability evaluation}

\begin{figure}[t!]
  \begin{center}
  \begin{tabular}{c|c|c|c}
 % \small
  %\hline
  wo & $\perp$ & Assign & CS \\
  \hline
   \includegraphics[width=0.09\textwidth]{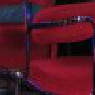} % \includegraphics[width=0.06\textwidth]{images/MIT_base_l2_casino/1_867_top_1.png} % \includegraphics[width=0.06\textwidth]{images/MIT_base_l2_casino/1_867_top_2.png}
   &  \includegraphics[width=0.09\textwidth]{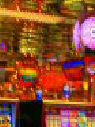} 
   & \includegraphics[width=0.09\textwidth]{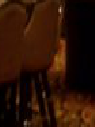}  %\includegraphics[width=0.06\textwidth]{images/MIT_base_l2_spars_casino/1_529_top_1.png}  
   & \includegraphics[width=0.09\textwidth]{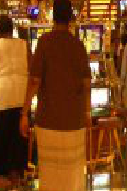}  %\includegraphics[width=0.06\textwidth]{images/MIT_discr_l2_casino/1_216_top_1.png}  \includegraphics[width=0.06\textwidth]{images/MIT_discr_l2_casino/1_216_top_2.png} 
   \\
     \includegraphics[width=0.09\textwidth]{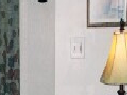}  %\includegraphics[width=0.06\textwidth]{images/MIT_base_l2_casino/2_1176_top_1.png}  \includegraphics[width=0.06\textwidth]{images/MIT_base_l2_casino/2_1176_top_2.png} 
     &  \includegraphics[width=0.09\textwidth]{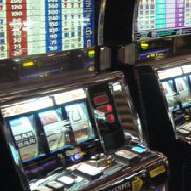} %\includegraphics[width=0.06\textwidth]{images/MIT_base_l2_ortho_casino/2_1060_top_1.png} \includegraphics[width=0.06\textwidth]{images/MIT_base_l2_ortho_casino/2_1060_top_2.png} 
     & \includegraphics[width=0.09\textwidth]{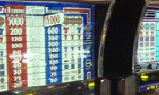}  %\includegraphics[width=0.06\textwidth]{images/MIT_base_l2_spars_casino/2_1199_top_1.png}  \includegraphics[width=0.06\textwidth]{images/MIT_base_l2_spars_casino/2_1199_top_2.png} 
     & \includegraphics[width=0.09\textwidth]{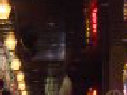}  %\includegraphics[width=0.06\textwidth]{images/MIT_discr_l2_casino/2_209_top_1.png}  \includegraphics[width=0.06\textwidth]{images/MIT_discr_l2_casino/2_209_top_2.png} 
     \\
    \includegraphics[width=0.09\textwidth]{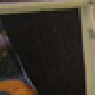}  %\includegraphics[width=0.06\textwidth]{images/MIT_base_l2_casino/3_689_top_1.png}  \includegraphics[width=0.06\textwidth]{images/MIT_base_l2_casino/3_689_top_2.png} 
    &  \includegraphics[width=0.09\textwidth,height=0.13\textwidth
    ]{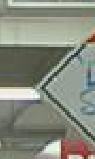} %\includegraphics[width=0.06\textwidth]{images/MIT_base_l2_ortho_casino/3_644_top_1.png} \includegraphics[width=0.06\textwidth]{images/MIT_base_l2_ortho_casino/3_644_top_2.png} 
    & \includegraphics[width=0.09\textwidth]{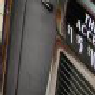}  %\includegraphics[width=0.06\textwidth]{images/MIT_base_l2_spars_casino/3_1107_top_1.png}  \includegraphics[width=0.06\textwidth]{images/MIT_base_l2_spars_casino/3_1107_top_2.png} 
    & \includegraphics[width=0.09\textwidth]{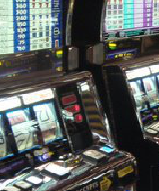}  %\includegraphics[width=0.06\textwidth]{images/MIT_discr_l2_casino/3_217_top_1.png}  \includegraphics[width=0.06\textwidth]{images/MIT_discr_l2_casino/3_217_top_2.png} 
    \\
  \end{tabular}
  \caption{ Three most important parts for the class \texttt{Casino}.
  %\caption{ Most important parts for the classes \texttt{Casino}, \texttt{Children room}, and \texttt{Florist} (first, second, and third rows). Each cells row contains the top three regions of the top three most discriminative parts (one part per line). 
  }
  \vspace{-0.2cm}
    \label{visu_regions}

  \end{center}
\end{figure}

We illustrate how parts help interpreting the DNN decision. First, Figure~\ref{visu_regions} shows the three most discriminative parts for the "casino" class of the MIT dataset according to equation~\ref{dMAC}. %Parts are represented by a list of three regions from the training set that are the most activated for the corresponding class.
We observe that constraints offer parts that look better aligned with categories.
% (better viewed on screen displays). 
%For a more exhaustive analysis, we encourage the reader to browse parts/class correspondences and regions visualisation at the following url: \url{http://...}. 

\begin{figure}
\centering
%  \begin{center}
%  \begin{tabular}{llll}
  \includegraphics[height=0.085\textwidth]{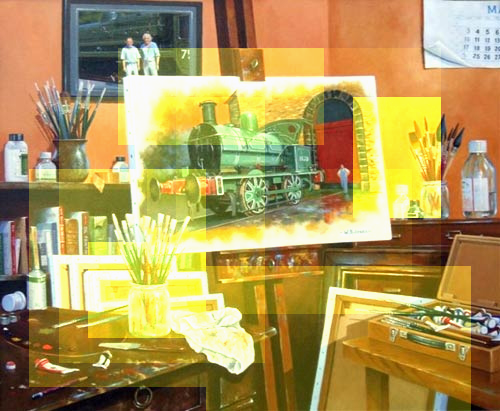}
  \includegraphics[height=0.085\textwidth]{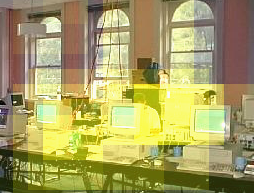} 
  \includegraphics[height=0.085\textwidth]{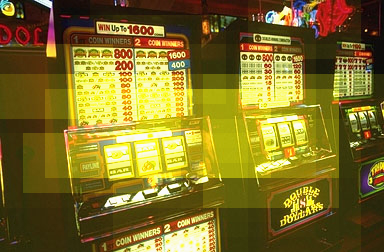}
  \includegraphics[height=0.085\textwidth]{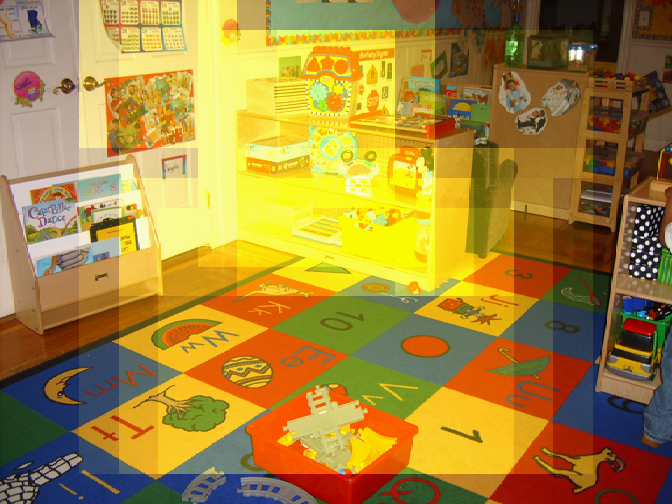}\\
%  \end{tabular}
 % \label{visu_heatmap}
 % \end{center}

\centering
\includegraphics[height =.08 \textwidth]{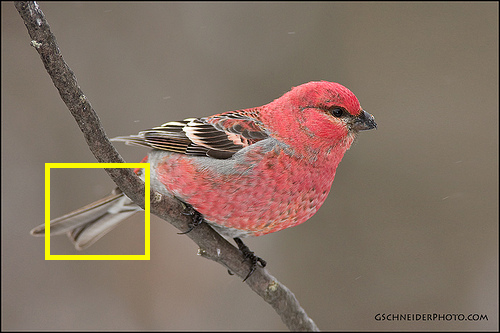}
\includegraphics[height =.08 \textwidth]{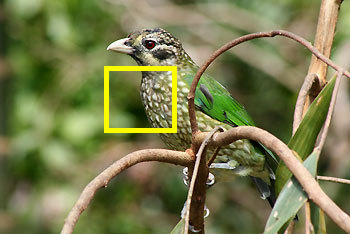}
\includegraphics[height =.08 \textwidth]{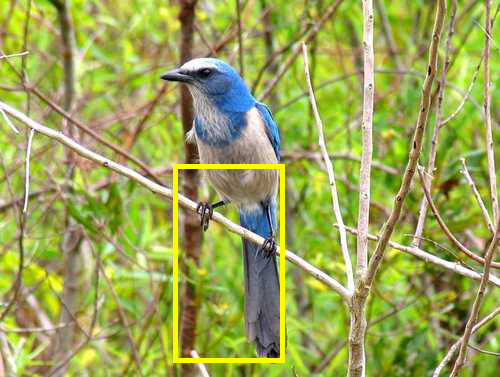}
\includegraphics[height =.08 \textwidth]{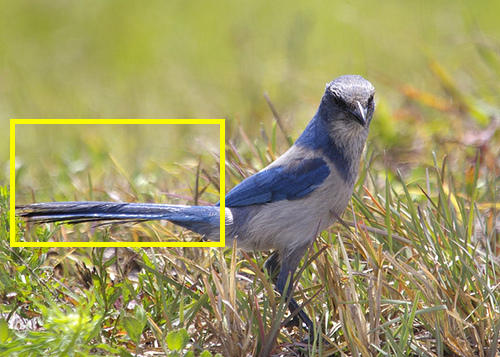}
\caption{
 Heatmap illustrations using test images of classes \texttt{Artstudio}, \texttt{Computer room}, \texttt{Casino} and \texttt{Kindergarden} on top row and most discriminant region used to classify birds test images on second row. 
}
  \vspace{-0.2cm}

\label{visu}
\end{figure}

\normalsize

%At inference time, the model makes its prediction based on regions that activate the most discriminative parts for a predicted class. We show various useful visualizations to explain decisions. 
Figure~\ref{visu} shows examples of heatmaps, where the most discriminative regions are highlighted in yellow. We note that the model focuses on semantically meaningful regions to take its decision. 
%
%Figure~\ref{visu_box} illustrates the potential of our method for explaining fine-grained classification, on the Birds dataset (CUB-100). 
Figure~\ref{visu} also highlights the most discriminative region used by the model to classify birds species. 
It is interesting to see that our model finds distinctive characteristics of species to recognize them: the long tail of the Florida Jay or the grainy appearance of the Spotted Catbird's chest.

\section{Conclusion}

%///Piste pour ameliorer les chiffres: data augmentation, RPN, finetuning of the backbone.

%/////////////////REDO

This paper presents DP-Net, an approach using a pretrained CNN to learn interpretable part representations.
The neural architecture %relying on part learning and detection, 
is accurate, scalable, and can deal with a variety of image recognition problems.
We introduce a number of constraints to drive part learning to favour interpretability of the model decisions.
Our experiments show interesting performance compared to standard global representations systems, across several networks and multiple datasets. We also provide evidence of the interpretability capabilities of our model by visualizing parts,  their  discriminative capabilities and their contribution in the decision.

%-------------------------------------------------------------------------
\bibliographystyle{IEEEbib}
\bibliography{refs}

\end{document}